\documentclass{article}

\usepackage{microtype}
\usepackage{graphicx}
\usepackage{subfigure}
\usepackage{booktabs} % for professional tables
\usepackage{multirow}
\usepackage{textcomp}
\usepackage{caption}
\usepackage{lscape}
\usepackage{rotating}
\usepackage{bm}
\usepackage[super]{nth}

\usepackage{hyperref}
\usepackage{amsmath}

\DeclareMathOperator*{\argmin}{arg\,min}

\usepackage[accepted]{icml2020}

\icmltitlerunning{Uncovering Coresets for Classification With MOEAs}

\begin{document}

\twocolumn[
\icmltitle{Uncovering Coresets for Classification\\With Multi-Objective Evolutionary Algorithms}



\icmlsetsymbol{equal}{*}

\begin{icmlauthorlist}
\icmlauthor{Pietro Barbiero}{polito}
\icmlauthor{Giovanni Squillero}{polito}
\icmlauthor{Alberto Tonda}{inrae}
\end{icmlauthorlist}

\icmlaffiliation{polito}{DAUIN, Politecnico di Torino, Torino, Italy}
\icmlaffiliation{inrae}{UMR 518 MIA, INRAE, Universit\'{e} Paris-Saclay, France}

\icmlcorrespondingauthor{Pietro Barbiero}{pietro.barbiero@polito.it}

\icmlkeywords{Machine Learning, Coresets}

\vskip 0.3in
]



\printAffiliationsAndNotice{\icmlEqualContribution} % otherwise use the standard text.

\begin{abstract}
%A \emph{coreset} is defined as a subset of the training set, using which a machine learning algorithm obtains performances similar to what it would deliver if trained over the whole original data. Advantages of coresets include improving training speed for the algorithms and easing human understanding of the dataset. Coreset discovery is an active and open line of research, as there is an obvious trade-off between limiting training size and quality of the results. Building on previous works, a novel evolutionary approach to coreset discovery is presented: starting from a population seeded with candidate coresets, a multi-objective evolutionary algorithm is set to modify them and eventually create new ones, to minimize both number of points in the set and classification error. Experimental results on non-trivial benchmarks show that the proposed approach is able to deliver results that allow a classifier to obtain lower error and/or better ability of generalizing on unseen data than state-of-the-art coreset discovery techniques. 

A \emph{coreset} is a subset of the training set, using which a machine learning algorithm obtains performances similar to what it would deliver if trained over the whole original data. Coreset discovery is an active and open line of research as it allows improving training speed for the algorithms and may help human understanding the results. Building on previous works, a novel approach is presented: candidate corsets are iteratively optimized, adding and removing samples. As there is an obvious trade-off between limiting training size and quality of the results, a multi-objective evolutionary algorithm is used to minimize simultaneously the number of points in the set and the classification error. Experimental results on non-trivial benchmarks show that the proposed approach is able to deliver results that allow a classifier to obtain lower error and better ability of generalizing on unseen data than state-of-the-art coreset discovery techniques. 
\end{abstract}

% \section{TODO}

% {\color{red}
% \begin{itemize}
%     \item Mettere in grassetto i risultati migliori (specificando quando non sono statisticamente separabili)
%     \item Commentare i risultati degli esperimenti (e.g. in micro-mass, siamo dell'1\% circa peggiori sul test, ma abbiamo core sets che sono il 25\% degli altri)
%     \item Siamo molto efficaci nei dataset grossi
%   \item Specificare quali sono le innovazioni rispetto al passato, specialmente il fatto che riusciamo ad affrontare dataset enormi
%     \item Se ci sta, mostrare il flow di un algoritmo evolutivo
%     \item Nelle figure, riusciamo a tratteggiare come la nostra soluzione "domini" le altre?
% \end{itemize}
% }

\section{Introduction}

A \emph{coreset} is like a small set of paradigmatic examples: a concept can be more effectively explained to a learner resorting to them than by enumerating a longer list of cases. Such an analogy, however, should not be pushed too far: a coreset in Machine Learning (ML) is more formally defined as the minimal set of training samples that allows a supervised algorithm to deliver a result as good as the one obtained when the whole set is used \cite{bachem2017practical}. Traditionally, the process of discovering a coreset consists in pinpointing the minimal subset of the data that is sufficient to achieve such good performances. The definition does not specify what is the task of the algorithm (classification, regression, or other), nor what is the quantitative measure used to evaluate its performances. The relevance of coresets is manifold: reducing the size of the training set can boost the performance of training, limit the memory requirements; and pinpointing the key elements required may provide an insight on the internal process, helping to explain, if not understand, ML results. 

While remarkable contributions to the problem date back to 1960s \cite{Efroymson1960}, discovering coresets for a specific ML task is still an open and active research line. Plainly, reducing the number of samples could impair the performance of the ML algorithm, and sometimes even state-of-the-art methodologies may not be able to preserve all the information of the original training set. However, coresets have so many advantages that the reduced quality might be acceptable. Therefore, the problem should be more productively framed as \emph{multi-objective}: minimize the size of the corset and maximize the quality of the final result, and eventually let the user to pinpoint the best compromise of these two conflicting goals. Moreover, as ML algorithms employ different techniques to accomplish the same goals, it is also reasonable to assume that they would need coresets of different size and shape to operate at the best of their possibilities. Starting from these two assumptions, and building on previous works~\cite{barbiero2019fundamental,barbiero2019making,barbiero2019evolutionary}, an Evolutionary Algorithm (EA) for coreset discovery is proposed.
% In this paper we face the problem of coreset from a different angle: instead of selecting the most suitable subset of the training data, we propose to build and optimize a set of virtual samples able to drive the ML algorithm in an optimal way. Using a metaphor as in the beginning of the section, we shift from the discovery of paradigmatic examples to the creation of prototypes. 

% what is different
When compared to similar works in literature \cite{barbiero2019evolutionary}, the proposed EA improves the state of the art as follows:
\begin{itemize}
    \item it exploits a new representation of coreset candidates in the EA, making it possible to tackle datasets of larger size;
    \item it does not require parameter tuning, as all relevant parameters are empirically derived from the size of the dataset.
\end{itemize}

The proposed approach exploits an Evolutionary Algorithm (EA) to drive the automatic selection of coresets. Preliminary experiments suggest that classifiers trained with such evolved samples are able to generalize better than those trained with the whole set. The rest of paper is organized as follows: a few necessary background elements are reported in Section \ref{sec:background}, the proposed approach is outlined in Section \ref{sec:alberto:proposed_approach}, while Section \ref{sec:experiments} reports the experimental evaluation, and Section \ref{sec:conclusions} concludes the paper.

\section{Background}
\label{sec:background}

\begin{figure*}[!htb]
    \centering
    \includegraphics[width=0.95\textwidth, trim=4.5cm 0cm 4.3cm 0cm, clip=true]{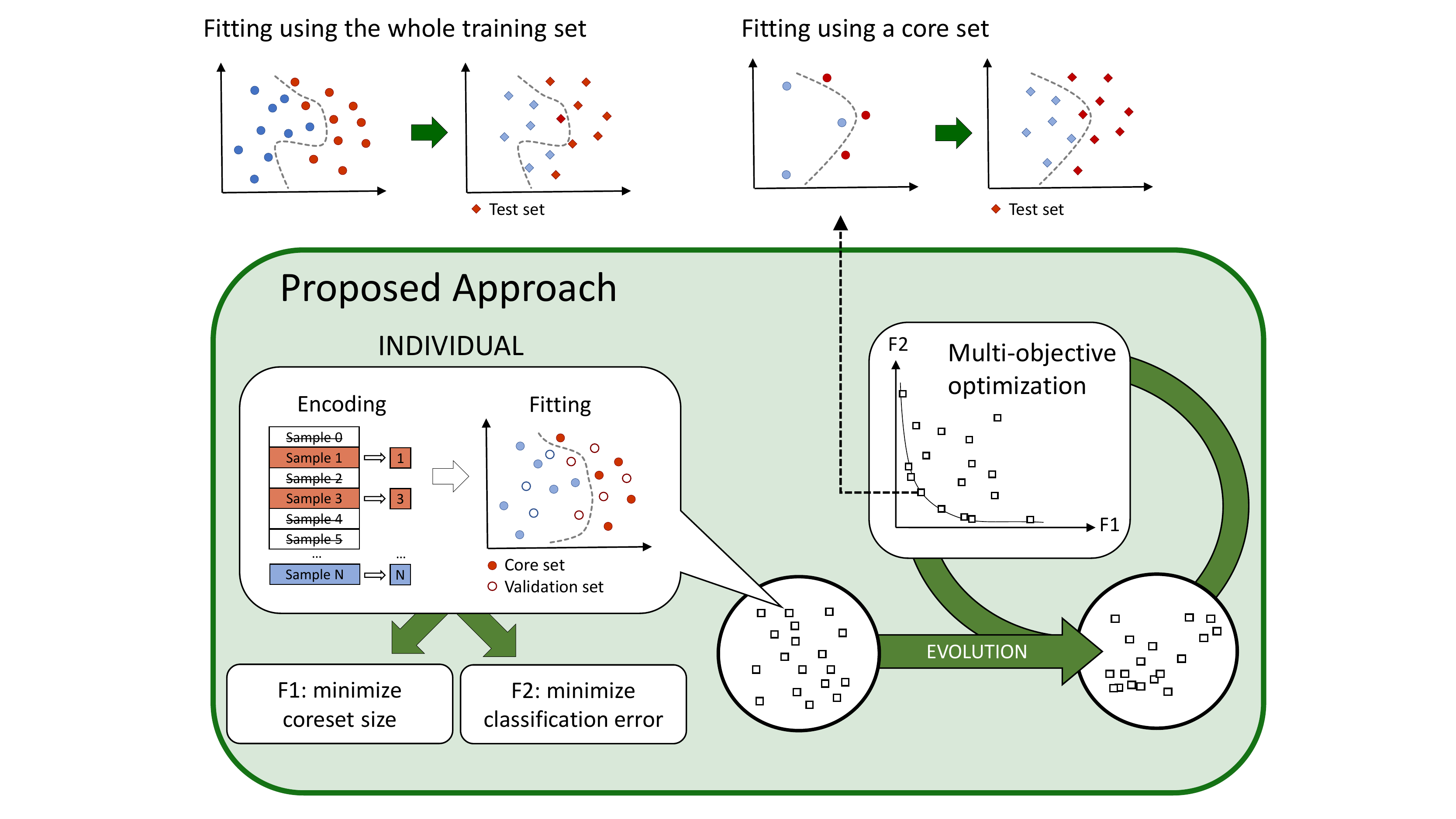}
    \caption{Scheme of the proposed approach. Given a dataset, split into training and test, the original training set is used to seed the initial population with coresets, consisting of sets training samples. Candidate solutions are evaluated on compactness (number of samples) and error (classifier trained on a candidate solution, tested on the original training set). New solutions are obtained through evolutionary operators. Once the evolution is complete, the coresets on the Pareto front undergo a final evaluation, training the classifier that will then compute its error on the (unseen) original test set, to evaluate the generality of the approach.}
    \label{figure:algorithm}
\end{figure*}

ML literature reports a number of approaches for the identification of coresets in different scenarios, starting from Forward Stagewise published in the 1966 \cite{Efroymson1960} up to the Greedy Iterative Geodesic Ascent (GIGA) appeared in 2018 \cite{campbell2018bayesian}. Other remarkable contributions to this research include Matching Pursuit \cite{mallat1993}, Orthogonal Matching Pursuit (Ortho Pursuit) \cite{Pati1993}, Frank-Wolfe \cite{clarkson2010coresets}, and Least-angle regression (LAR) \cite{Efron2004, Boutsidis2013}. The original Frank-Wolfe algorithm applies in the context of maximizing a concave function within a feasible polytope by means of a local linear approximation. In Section~\ref{sec:experiments}, we refer to the Bayesian implementation of the Frank-Wolfe algorithm designed for core set discovery. This technique, described in~\cite{campbell2017automated}, aims to find a linear combination of approximated likelihoods (which depends on the core set samples) that is similar to the full likelihood as much as possible. GIGA is a greedy algorithm that further improves Frank-Wolfe. In~\cite{campbell2018bayesian}, the authors show that computing the residual error between the full and the approximated likelihoods by using a geodesic alignment guarantees a lower upper bound to the error at the same computational cost. On the other hand, Forward Stagewise, Least-angle regression, Matching Pursuit and Orthogonal Matching Pursuit were all originally devised as greedy algorithms for dimensionality reduction, but have been later applied to coreset discovery, as this last problem represents the transpose of feature selection, choosing samples instead of features. The simplest of the group is Forward Stagewise, which projects high-dimensional data in a lower dimensional space by selecting, one at a time, the features whose inclusion in the model gives the most statistically significant improvement. Matching Pursuit, on the other hand, includes features having the highest inner product with a target signal, while its improved version Orthogonal Matching Pursuit at each step carries an orthogonal projection out. Similarly, Least-angle regression increases the weight of each feature in the direction equiangular to its correlations with the target signal.

While all the above approaches take the size of the coreset as an input of the problem, recently multi-objective EAs were proposed to determine the best trade-off between final performances and coreset size \cite{barbiero2019fundamental,barbiero2019evolutionary}.
% \cite{barbiero2019fundamental}. 
Since the 1990s, Evolutionary Computation (EC) techniques have been used to optimize ML frameworks, demonstrating the possibility to determine semi-optimal topologies or to efficiently train artificial neural networks \cite{angeline1994evolutionary,maniezzo1994genetic,frean1990upstart}. A topic where the use of EC soon appeared promising is feature selection \cite{vafaie1992genetic,kim2000feature}: as the performance of ML algorithms are quite sensitive to the choice of the features, a smart selection procedure is quite beneficial, and EC allowed to automatize the procedure. Selecting features using the eventual performance of a ML algorithm as fitness function is indisputably related to the problem of selecting training-set elements with the same goal.

EAs are stochastic optimization techniques, loosely inspired by the neo-Darwinian paradigm of natural selection. Candidate solutions (\emph{individuals}) are encoded in appropriate data structures (\emph{genomes}); the algorithm manipulates a set of them (\emph{population}), trying to generate better ones. An evaluation function (\emph{fitness function}) is used to assess the extent to which a individual solves the problem; such value controls the probability that the individual is selected for reproduction and survival. In each discrete step of the algorithm  (\emph{generation}), new individual (\emph{offspring}) are first generated using \emph{recombination} and \emph{mutation} (cumulatively called \emph{genetic operators}), then evaluated, and eventually the less fit are discarded. The EA stops when a user-defined threshold is reached, typically a limit on the number of generations. 

%{\color{red} Indeed, while ML and EC progressed almost independently in the past decades, the two fields are deeply linked. The \emph{obvious connection} between the processes of learning and evolution has been pointed out by Alan Turing back in 1950 \cite{turing2009computing} and the very term ``machine learning'' was coined in 1959 by Arthur Lee Samuel, a pioneer in the field of computational intelligence those attempts to devise a checkers player are frequently listed in EC histories \cite{samuel1959some}. Seminal works in EC explicitly refer to ML, far before the beginning of the current ML windfall \cite{goldbergholland1988genetic,goldberg1989genetic,grefenstette1993genetic, farmer1986immune}.}

Among the most successful applications of EC, multi-objective optimization often takes the center stage. Optimization problems with contrasting objectives have no single optimal solution. Each candidate represents a different compromise between the multiple conflicting aims. Yet, it is still possible to search for \emph{optimal trade-offs}, for which an objective cannot be improved without degrading the others. The set of such optimal compromises is called \emph{Pareto front} from the \nth{19} century Italian engineer Vilfredo Pareto. Multi-objective evolutionary algorithms (MOEA) currently represent the state of the art for problems with contradictory objectives, and are able to obtain good approximations of the true Pareto front in a reasonable amount of time. One of the most known MOEAs is the Non-Sorting Genetic Algorithm II (NSGA-II)~\cite{deb2002fast}, that makes use of a special mechanism to spread candidate solutions on the Pareto front as evenly as possible, with considerable efficiency for problems with few objectives.

Interestingly, conflicting objectives to optimize abound in ML, with models being trade-offs between fitting and complexity, and coresets being compromises between number of samples considered and quality of the final result.

\section{Proposed approach}
\label{sec:alberto:proposed_approach}

% Given the practical usefulness and limitations of coreset discovery, a novel alternative is proposed: prototype discovery.
In ML, coresets are defined as a subset of the initial (training) dataset that, used on its own, does not significantly impact the quality of the results of a ML algorithm with respect to the original dataset. In other terms, coresets can be seen as a \textit{summary} of the information contained in the original dataset, encompassing all the essential data to obtain the correct behavior of the ML algorithm. 

As the search space of all possible coresets of varying size for a given problem is clearly vast, it is necessary to resort to stochastic optimization to efficiently explore it. As said before, oreset discovery is inherently a multi-objective problem. Building on previous works on evolutionary coreset discovery~\cite{barbiero2019fundamental,barbiero2019evolutionary}, the proposed algorithm extends and improves the evolutionary approach making it suitable for tackling datasets of larger size without requiring parameter tuning. While the focus of this work is on coreset discovery for classification, it could easily be extended to regression problems. A summary of the proposed algorithm is presented in Figure~\ref{figure:algorithm}.

%The base of the MOEA used for the proposed approach is NSGA-II, employing the classical crowding mechanisms and using standard parameters, with the following exceptions {\color{red} which ones?}. 

The algorithm can be summarized as follows. The dataset $\mathcal{D}$ is split into three groups for training ($\mathcal{D}_t$, $90\%$ of samples), validating ($\mathcal{D}_v$, $10\%$ of samples), and testing ($\mathcal{D}_u$, $10\%$ of samples). Given the training sample $s_i = (\vec{x}^{i},y^{i}) \in \mathcal{D}_t$, where $\vec{x}^i$ is the feature vector of $s_i$ and $y^i$ the corresponding class label, respectively, the objective is to estimate the probability that it belongs to the coreset $\mathcal{C}_j$, given the ML model (i.e., the \textit{classifier}) $f$:
\begin{equation}
    p( s_i \in \mathcal{C}_j | f)
\end{equation}
Each coreset candidate $\mathcal{C}_j$ is then used to fit the model parameters:
\begin{equation}
    p( f_j | \mathcal{C}_j )
\end{equation}
Finally, the trained classifier is used to make inferences on the training set:
\begin{equation}
    p( y_t^{i} |  \vec{x}_t^{i}, f_j )
\end{equation}
The proposed approach makes it possible to evolve several solutions $\mathcal{C}_j$ approximating the coreset problem, reducing both the set size and the classification error $\eta$:
\begin{equation}
    \argmin_{\mathcal{C}_j}
    \begin{cases}
    |\mathcal{C}_j|\\
    \eta(f_j,\mathcal{D}_t)
    \end{cases}
\end{equation}
The classification error has been defined as $1$ minus the weighted $F_1$ score \cite{sorensen1948method,chinchor1991muc} to account for class imbalance in multi-class datasets:
\begin{equation}
    \eta = 1 - \frac{1}{\sum_{l \in L} |y_l|} \sum_{l \in L} |y_l| F_1 (\hat{y_l}, y_l)
\end{equation}
where $L$ is the set of labels, $\hat{y}$ the set of \textit{predicted} sample/label pairs, $y$ the set of \textit{true} sample/label pairs, $y_l$ the subset of $y$ with label $l$, and $F_1$ given by:
\begin{equation}
    F_1 = 2 \times \frac{precision \times recall}{precision + recall}
\end{equation}

At the end of the evolution, the validation set $\mathcal{D}_v$ is used in order to evaluate the final solutions along the Pareto front. The maximum likelihood estimation for the evolved solutions is given by the core set providing the best score on the validation set:
\begin{equation}
    \hat{\mathcal{C}}^{MLE} = \arg\max_{\mathcal{C}_j} 1-\eta(f_j,\mathcal{D}_v)
\end{equation}
Finally, $\hat{\mathcal{C}}^{MLE}$ is used to train the model:
\begin{equation}
    p( f_{MLE} | \hat{\mathcal{C}}^{MLE} )
\end{equation}
to make inferences on an unseen test set $\mathcal{D}_u$ and to compute the classification error:
\begin{equation}
    \eta(f_{MLE},\mathcal{D}_u)
\end{equation}

%{\color{red} ANALYSIS OF THE COMPLEXITY (TIME, SPACE, SAMPLESIZE); PROBABLY NOT NEEDED, AS NSGA-II IS ALREADY PUBLISHED}

\subsection{Genotype of a candidate solution}
\label{ssec:alberto:genotype_of_a_candidate_solution}
% genotype
The genotype of a candidate solution is represented by a list of integers encoding the position of core samples in the original dataset. The list size may vary for different candidate solutions according to the number of core samples selected.
% by a matrix, where each row encodes the real-valued features of one virtual data point, plus an integer that associates the virtual point to one of the known classes.

\subsection{Fitness functions}
\label{ssec:alberto:fitness_functions}
% fitness function
The two fitness functions used in this multi-objective problem are the size of the coreset (to be minimized), and the error of the target classifier trained on core samples, evaluated on the original dataset (to be minimized).

\subsection{Population initialization}
\label{ssec:alberto:population_initialization}
% initialization
The starting population of the MOEA is initialized with coresets of random size. The minimum size corresponds to the number of classes in the problem, so that each candidate solution has at least one data point associated to each class; the maximum size is defined as a 1/10 of the original dataset size. Data points in each of such coresets are randomly drawn from the original dataset.

\subsection{Evolutionary operators}
\label{ssec:alberto:evolutionary_operators}
% operators
When generating new candidate solutions, the MOEA randomly selects two individuals and applies the following operators: i. cross over the two candidate solutions, by randomly distributing core samples contained in both between two children solutions; ii. mutate the children solutions adding or removing one sample from their respective list; iii. repair children if the obtained representations violate feasibility constraints (e.g. minimum number of classes, maximum and minimum coreset size).
% the MOEA selects one of the following operators, with flat probability: i. mutate one feature of one archetype in an archetype set, using a Gaussian mutation with $\mu=0$ and $\sigma=0.1$; ii. mutate the class associated to one archetype, changing it to a different valid class in the problem; iii. remove one archetype from the archetype set, checking that the solution is still valid, containing at least one archetype per class; iv. add one random sample from the original training set to an archetype set; v. cross over two candidate solutions, by randomly distributing each archetype contained in both between two children solutions.

\subsection{Parameters setting}
\label{ssec:hyperparameters}
Most parameters in an EA can be derived from a single value, the population size $\mu$. In many practical applications, $\mu$ is set through a trial-and-error approach; but there are alternatives, for example the empirical formula used by the state-of-the-art single-objective optimizer Covariance Matrix Adaptation Evolution Strategy \cite{hansen2001completely}, where population size is determined starting from problem size.

We decided to follow a similar approach, and derive the parameters of the proposed algorithm from the size of the problem, that in the case of coreset discovery can be estimated as $2^k$, the total number of possible different coresets of size $[1,k]$, that represents the search space that the algorithm will need to explore in order to find the best coreset candidate. We fix $k=0.1*N$, where $N$ is the total number of samples in the considered data set, as in most practical cases desirable coresets are 10\% or less of the initial samples.

Following the empirical formulas presented in \cite{hansen2001completely}, starting from $k$ we can then derive the other necessary parameters for NSGA-II:

\begin{align}
    &\mu = \lfloor \max(100, log_{10} 2^k) \rfloor \\
    &\lambda = 2 * \mu\\
    &\mathcal{G} = \lfloor \max(100, log_{10} 2^{k\cdot0.5}) \rfloor
\end{align}

Where $\mu$, as mentioned above, is the size of the population; $\lambda$ is the size of the offspring, the number of new candidate solutions generated at each iteration; and $\mathcal{G}$ is the number of iterations after which the algorithm will stop. When compared to the empirical formulas of CMA-ES for deriving parameters, the main difference is that we increased the minimum size of the population and the minimum number of iterations, as the specific problem we are tackling is quite challenging.

% pop_size: int = 100, max_generations: int = 100, max_points_in_core_set: int = 100, n_splits: int = 10, random_state: int = 42

% k = int(0.1 * X.shape[0])

% self.max_generations_ = np.max([self.max_generations, int(math.log10(2**int(0.5 * k)))])
% self.pop_size_ = np.max([self.pop_size, int(math.log10(2**k))])
% self.offspring_size_ = 2 * self.pop_size_
% self.maximize_ = True
% self.individuals_ = []
% self.scorer_ = get_scorer(self.scoring)
% self.max_points_in_core_set_ = np.max([k, self.max_points_in_core_set])

\section{Experimental Evaluation}
\label{sec:experiments}

All the necessary code for the experiments has been implemented in Python 3, relying upon open-source libraries as \texttt{scikit-learn v0.22}\footnote{scikit-learn: Machine Learning in Python, \url{http://scikit-learn.org/}}~\cite{scikit-learn} and \texttt{inspyred v1.0}\footnote{inspyred: Bio-inspired Algorithms in Python, \url{https://pythonhosted.org/inspyred/}}~\cite{inspyred}, and \texttt{bayesiancoresets v0.8}\footnote{bayesiancoresets: Coresets for approximate Bayesian inference, \url{https://github.com/trevorcampbell/bayesian-coresets/}}~\cite{campbell2019sparse}. The code is freely available under GNU Public License from a GitHub public repository\footnote{GitHub, \url{https://github.com/albertotonda/prototype-set-discovery}}. NSGA-II is used with the default parameters set by \texttt{inspyred}, with the exceptions described in subsection \ref{ssec:hyperparameters}. In order to generate reproducible results, all algorithms that exploit pseudo-random elements in their training process have been set with a fixed seed. Before running coreset selection algorithms, each dataset has been standardized removing the mean and scaling to unit variance (\texttt{StandardScaler}~\cite{zill2011advanced}). All the experiments have been run on the same machine: nn AMD EPYC 7301 16-Core Processor at 2 GHz equipped with 64 MiB memory.

Table \ref{tab:data-sets} summarizes the main characteristics of the benchmark datasets used for the experiments. All of them have been downloaded from the OpenML public repository~\cite{OpenML2013,OpenMLPython2019}.

% \usepackage{booktabs}
% \usepackage{graphicx}
\begin{table}[!htb]
\centering
\caption{Benchmark datasets}
\vskip 0.1in
\label{tab:data-sets}
\begin{center}
\begin{small}
\begin{sc}
% \resizebox{\textwidth}{!}{%
\begin{tabular}{@{}lrrrr@{}}
\toprule
data set & samples & features & classes & NaNs \\ \midrule
micro-mass & 571 & 1,301 & 20 & 0 \\
soybean & 683 & 36 & 19 & 2,337 \\
credit-g & 1,000 & 21 & 2 & 0 \\
kr-vs-kp & 3,196 & 37 & 2 & 0 \\
abalone & 4,177 & 9 & 28 & 0 \\
isolet & 7,797 & 618 & 26 & 0 \\
jm1 & 10,885 & 22 & 2 & 25 \\
gas-drift & 13,910 & 129 & 6 & 0 \\
mozilla4 & 15,545 & 6 & 2 & 0 \\
letter & 20,000 & 17 & 26 & 0 \\
Amazon & 32,769 & 10 & 2 & 0 \\
electricity & 45,312 & 9 & 2 & 0 \\
mnist & 70,000 & 785 & 10 & 0 \\
covertype & 581,012 & 55 & 17 & 0 \\ \bottomrule
\end{tabular}%
% }
\end{sc}
\end{small}
\end{center}
\vskip -0.1in
\end{table}

\begin{figure*}[!htb]
    \centering
    
    \includegraphics[width=0.48\textwidth]{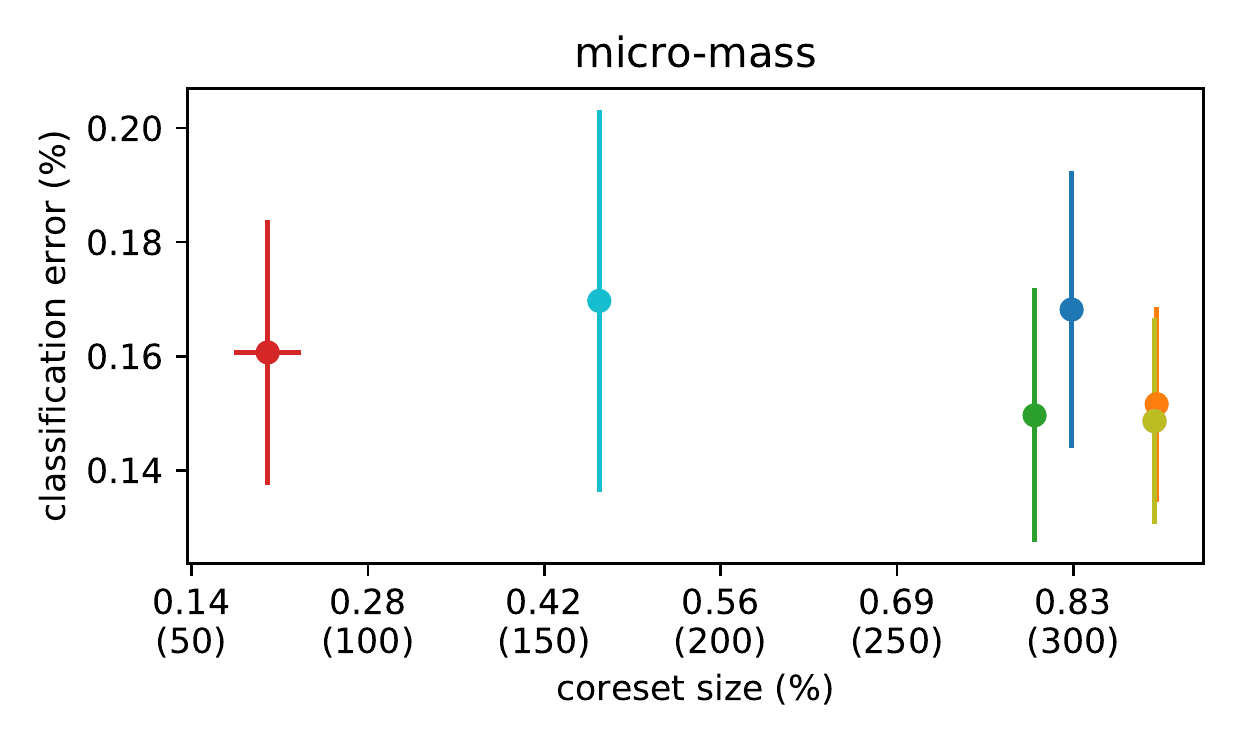}
    \includegraphics[width=0.48\textwidth]{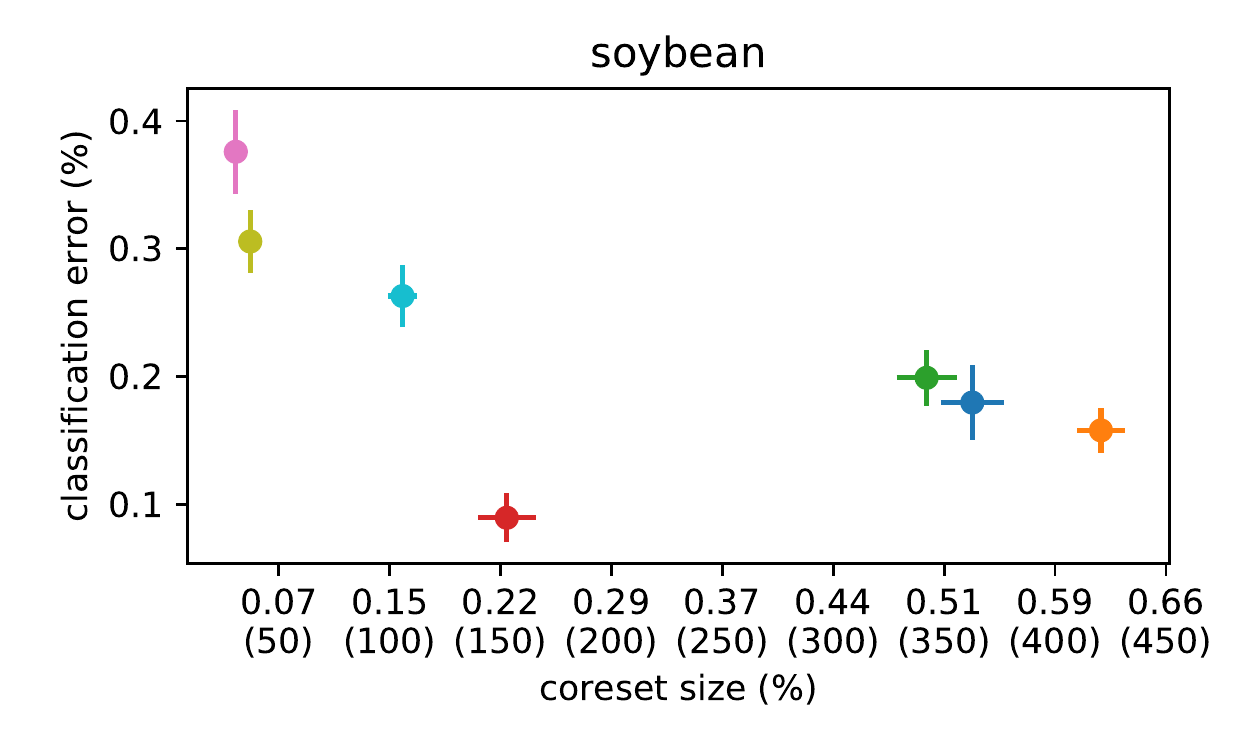}\\
    \includegraphics[width=0.48\textwidth]{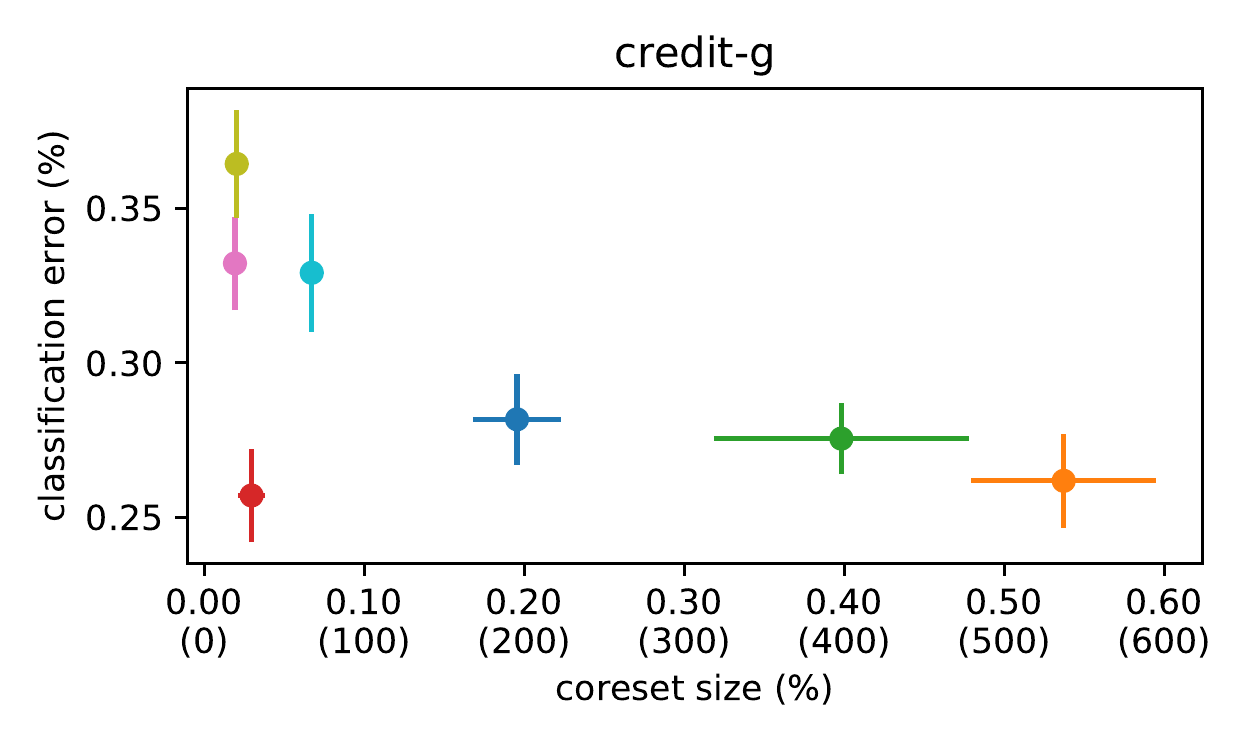}
    \includegraphics[width=0.48\textwidth]{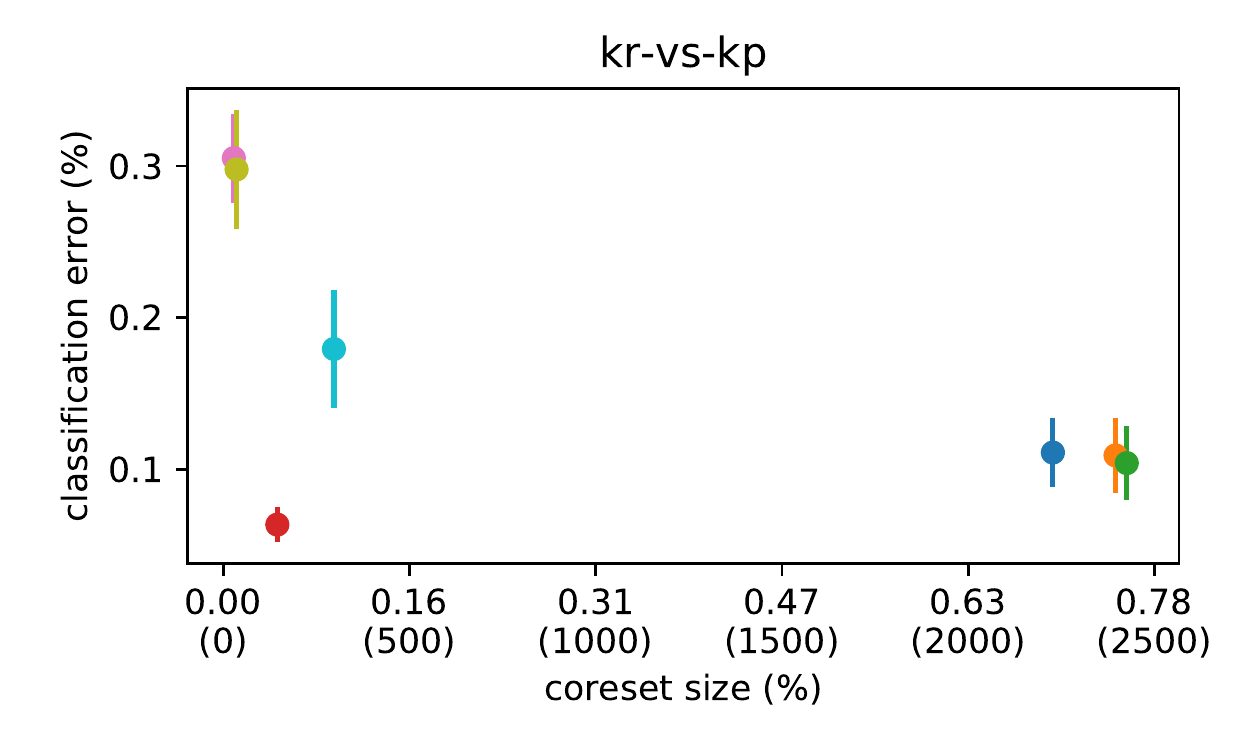}\\
    \includegraphics[width=0.48\textwidth]{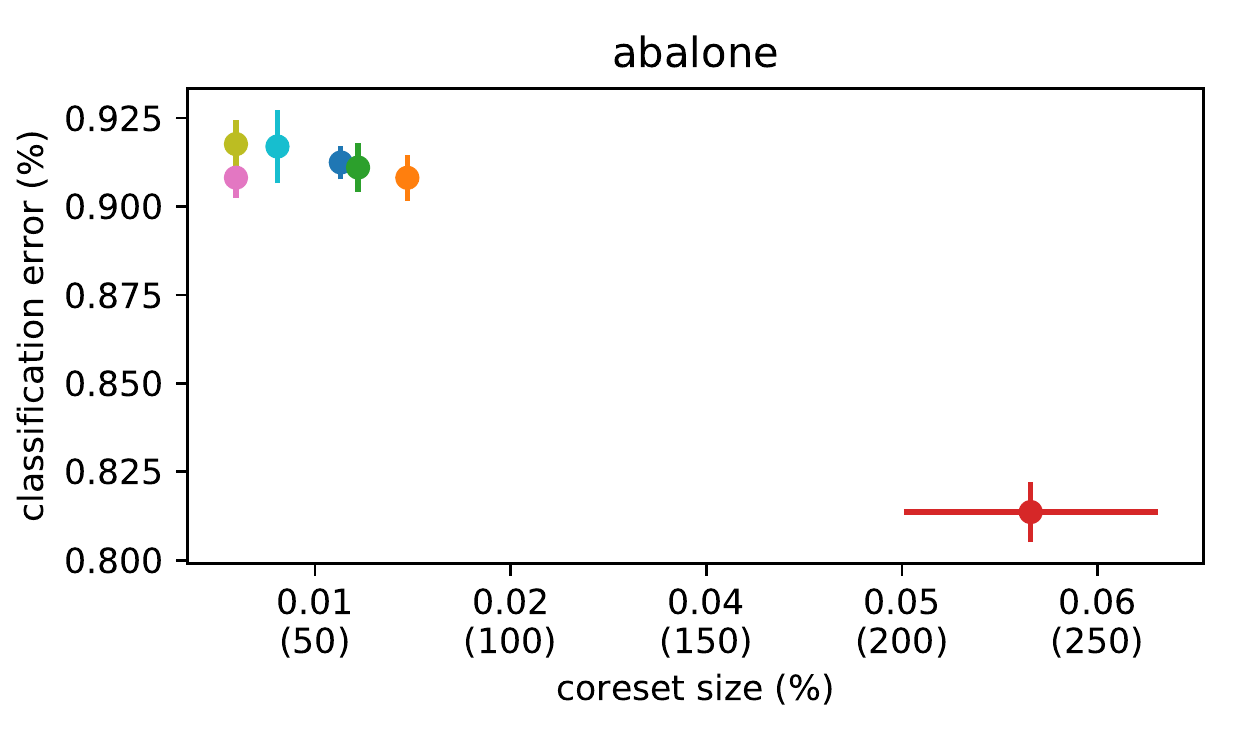}
    \includegraphics[width=0.48\textwidth]{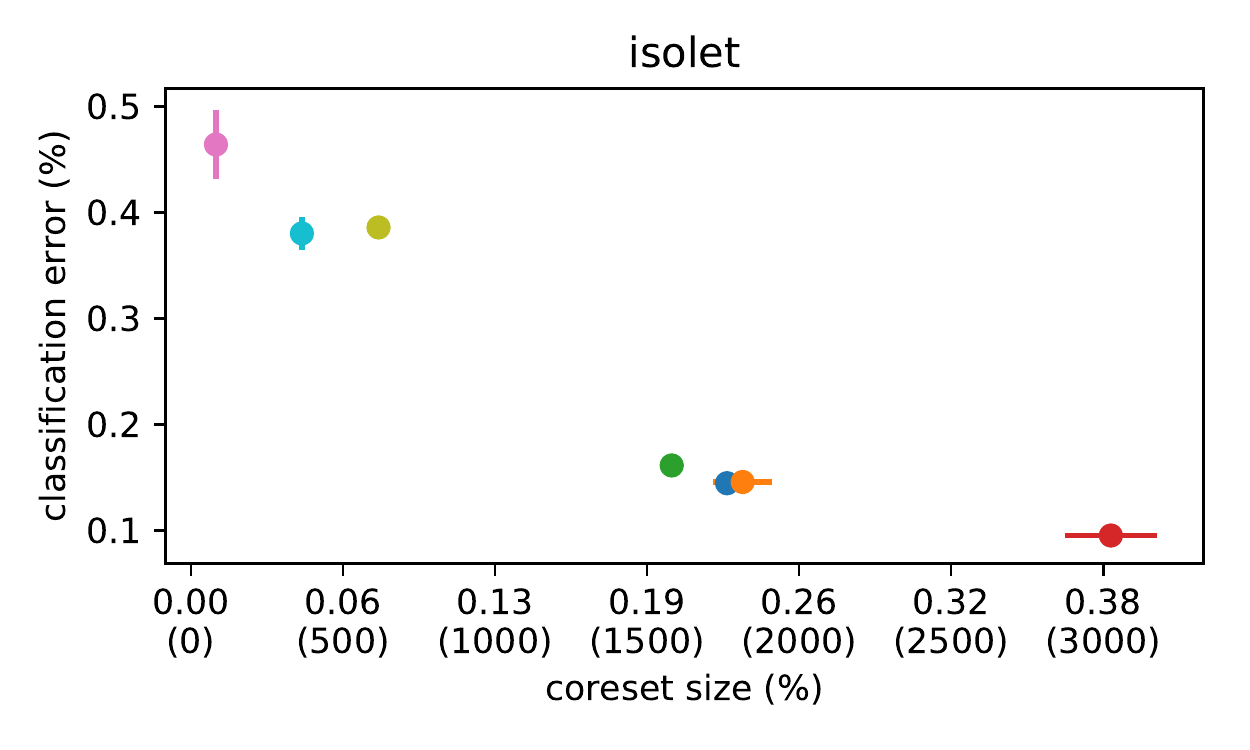}\\    
    \includegraphics[width=0.48\textwidth]{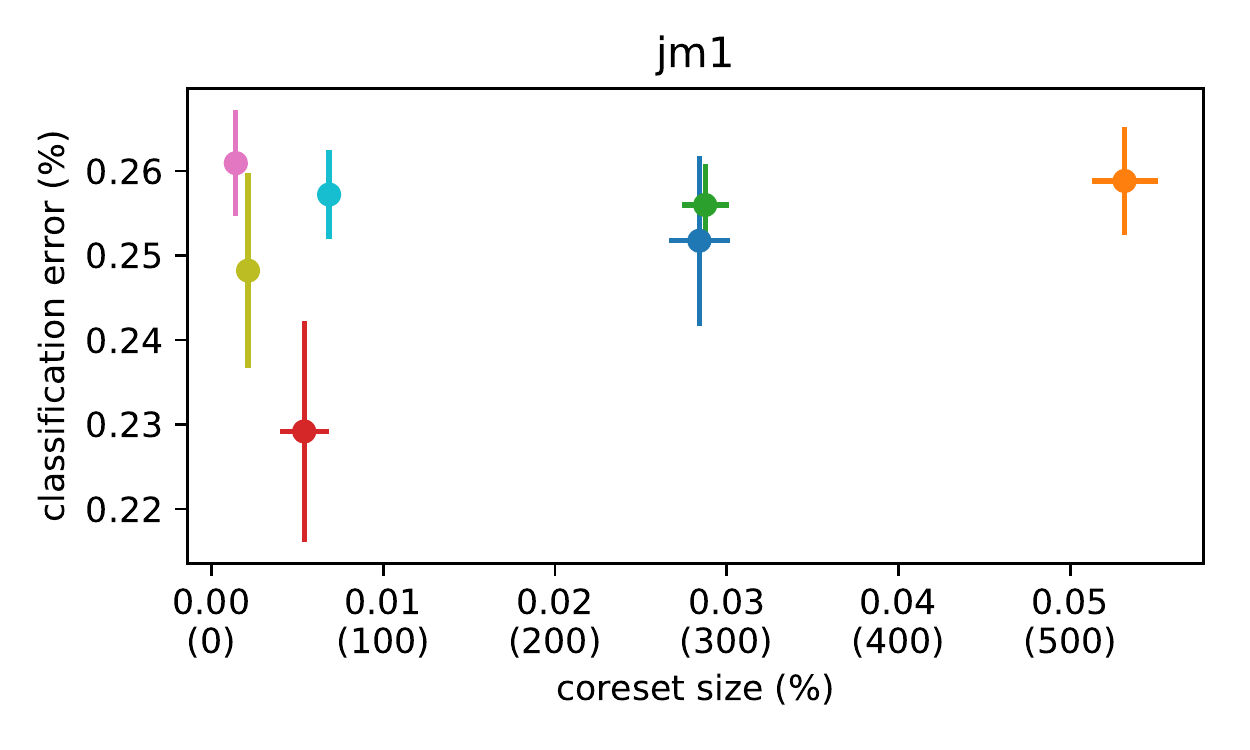}
    \includegraphics[width=0.48\textwidth]{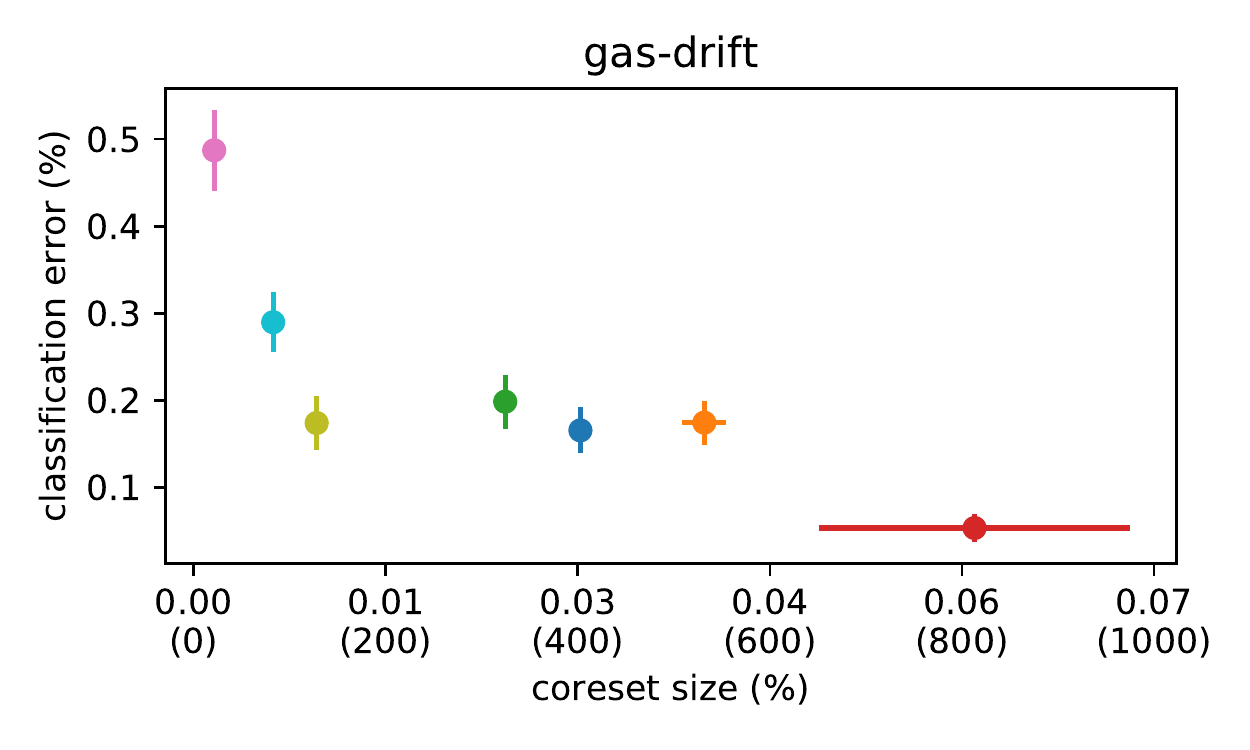}
    \includegraphics[width=0.8\textwidth, trim=1.5cm 1cm 1.5cm 2cm, clip=true]{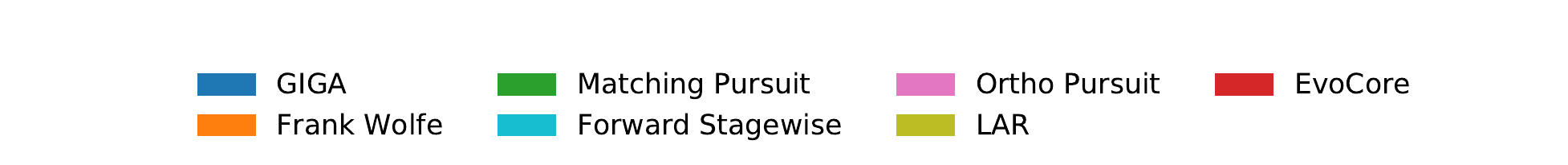}

    \caption{The figures show for each benchmark dataset the quality of extracted coresets as a function of coreset size (lower is better), and classification error obtained using an instance of the \texttt{Ridge} classifier (lower is better). The error bars represent the standard error of the mean.}
    \label{fig:exp1}
    
\end{figure*}

\begin{figure*}[!htb]
    \centering

    \includegraphics[width=0.48\textwidth]{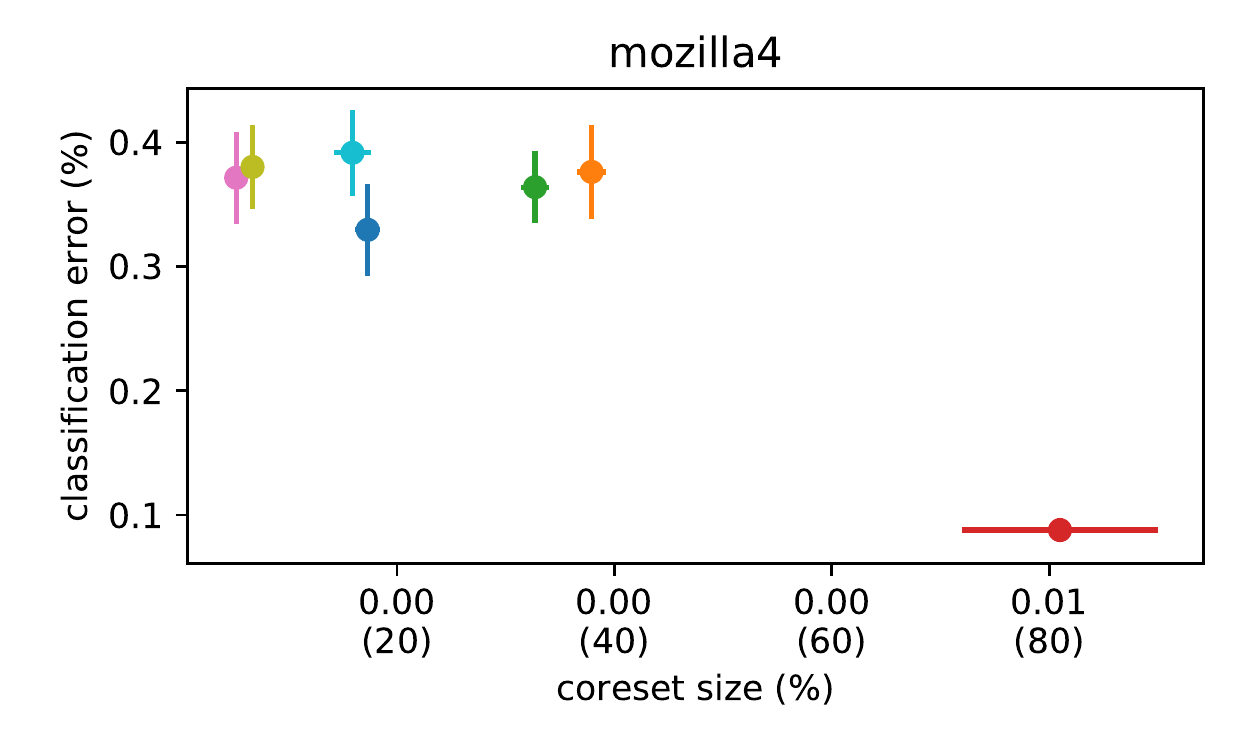}
    \includegraphics[width=0.48\textwidth]{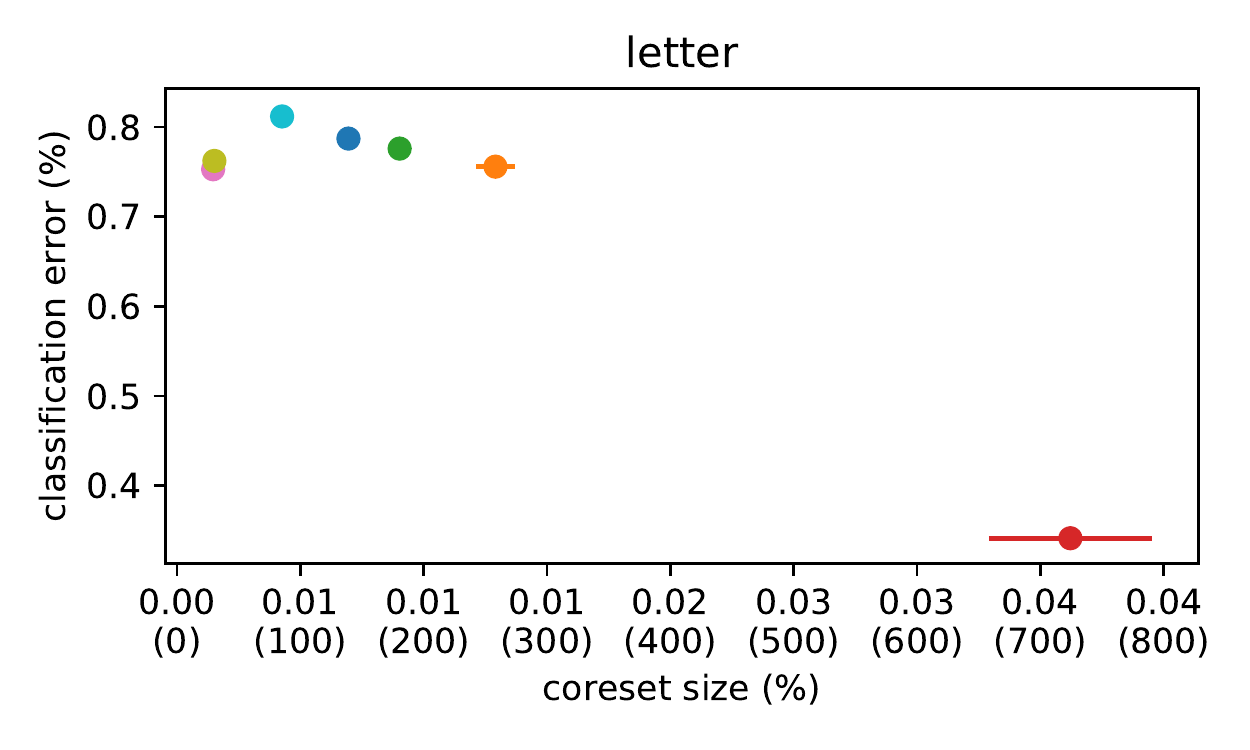}\\
    \includegraphics[width=0.48\textwidth]{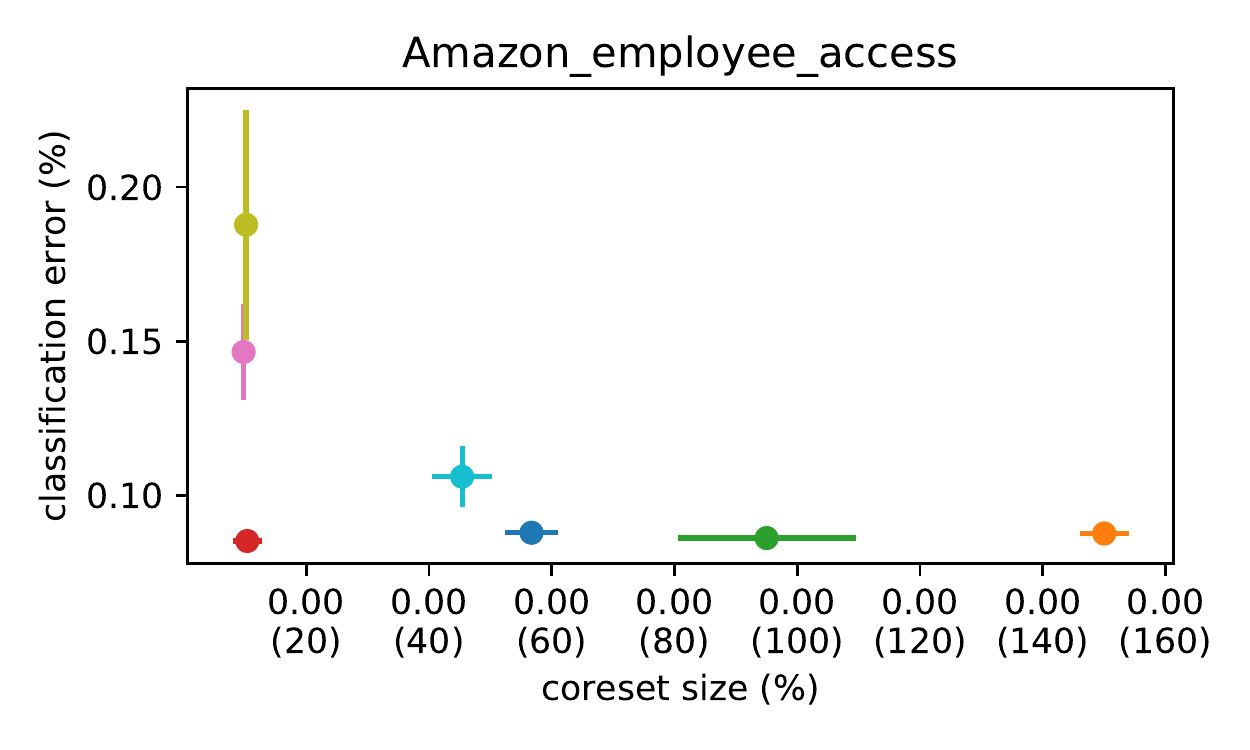}
    \includegraphics[width=0.48\textwidth]{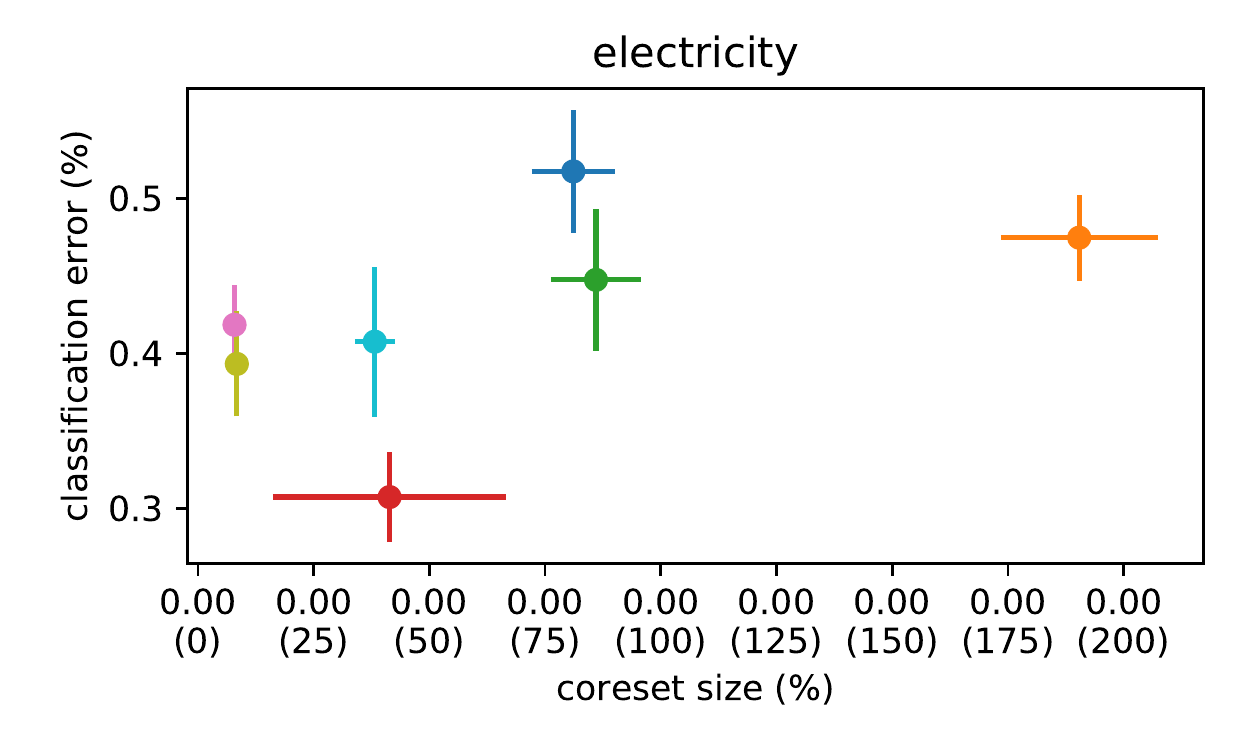}\\    \includegraphics[width=0.48\textwidth]{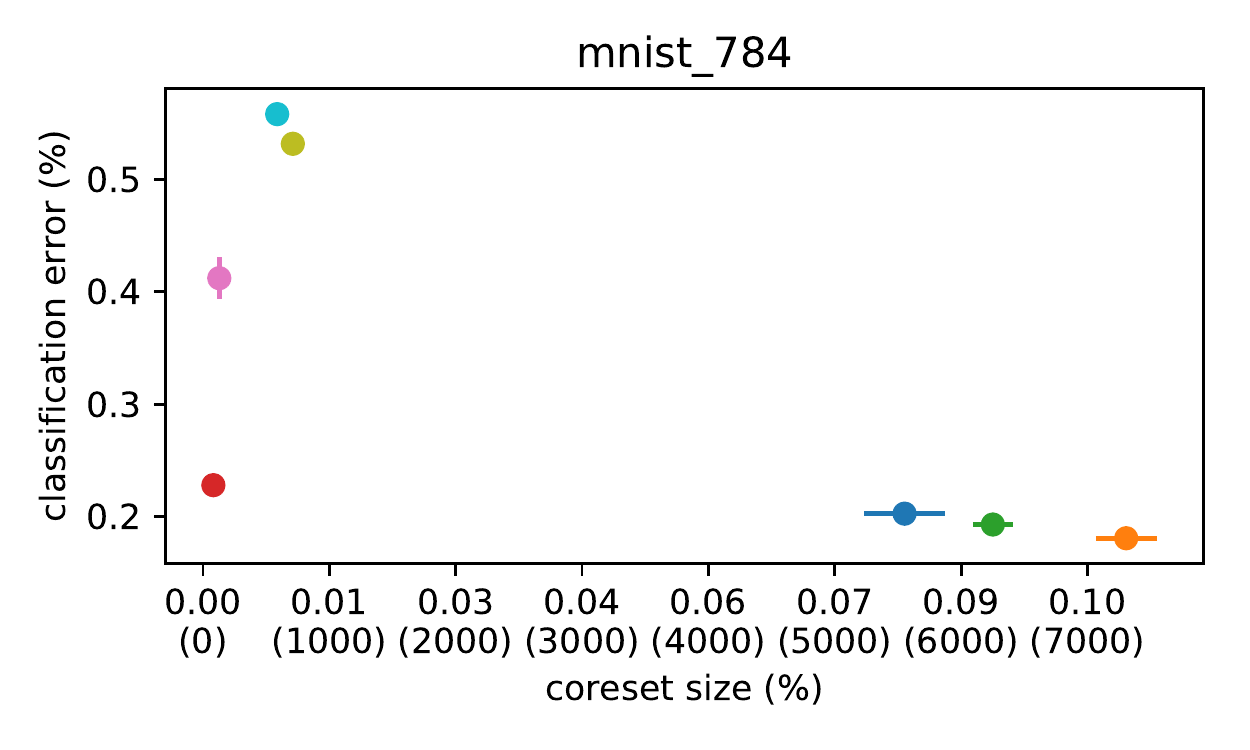}
    \includegraphics[width=0.48\textwidth]{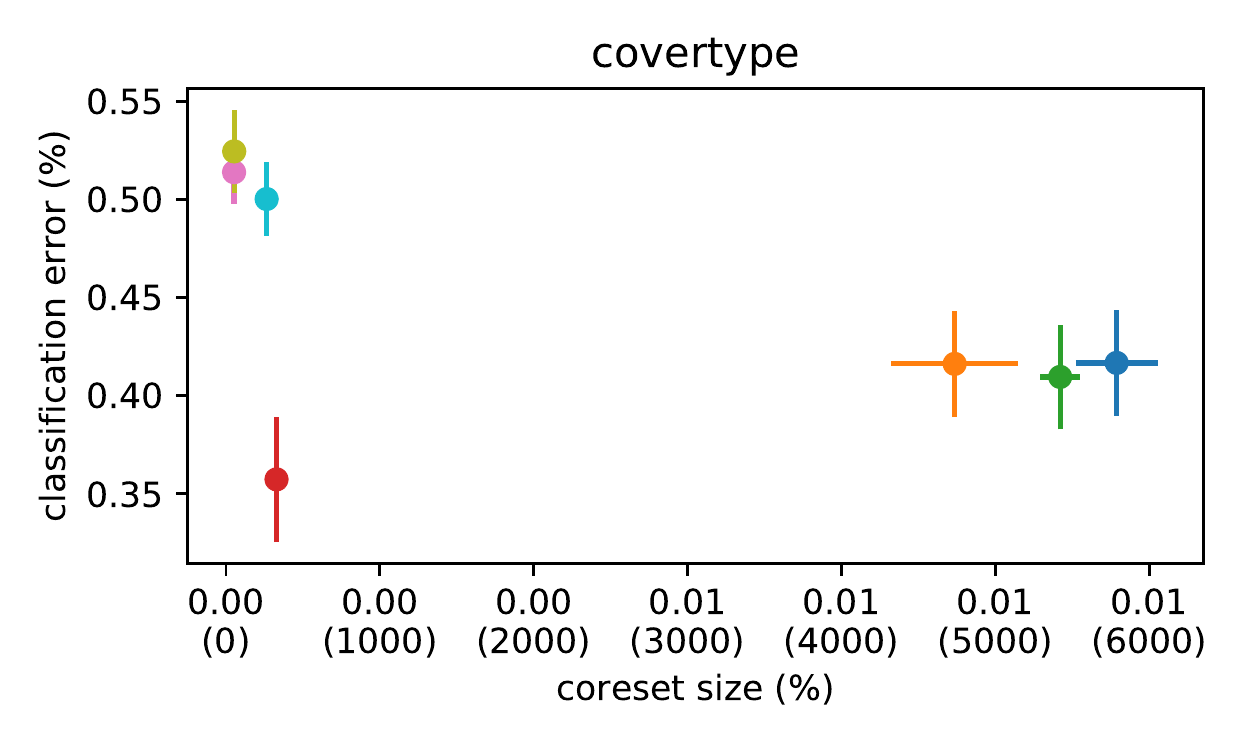}
    \includegraphics[width=0.8\textwidth, trim=1.5cm 1cm 1.5cm 2cm, clip=true]{figs/legend.pdf}

    \caption{The figures show for each benchmark dataset the quality of extracted coresets as a function of coreset size (lower is better), and classification error obtained using an instance of the \texttt{Ridge} classifier (lower is better). The error bars represent the standard error of the mean.}
    \label{fig:exp2}
    
\end{figure*}

The proposed approach has been tested over a $10$-fold cross-validation against state-of-the-art coreset discovery algorithms: GIGA~\cite{campbell2018bayesian}, Frank-Wolfe~\cite{clarkson2010coresets}, Matching Pursuit~\cite{Pati1993}, Orthogonal Mathcing Pursuit~\cite{Pati1993}, LAR~\cite{Efron2004,Boutsidis2013}, and Forward Stagewise~\cite{Efroymson1960}.
% , and random sampling (used as a baseline method). 
For each fold, coresets have been extracted and used to train an instance of the \texttt{Ridge}~\cite{Tikhonov1943} classifier. This classifier has been chosen both for its high train speed and its generalization ability in a variety of experimental settings. Figures \ref{fig:exp1} and \ref{fig:exp2} show for each benchmark dataset the quality of extracted coresets as a function of coreset size (lower is better), and classification error (lower is better). The coreset size has been reported both as the absolute number of core samples (in round brackets) and as the ratio
$|\hat{\mathcal{C}}^{MLE}| / |\mathcal{D}_t|$ (number of core samples over the total number of training samples).
% $\# \text{core samples} / \# \text{dataset samples}$. 
The error bars represent the standard error of the mean.
% For most of the benchmark datasets EvoCore was able to find better compromises between the two objectives, supporting the hypothesis that coreset discovery is indeed a multi-objective problem. 
Details about experimental results are shown in the Table \ref{tab:results1} and \ref{tab:results2}. 
The results can be summarized as follows:
\begin{itemize}
    \item Concerning just mean classification error, EvoCore outperformed state-of-the-art techniques on all datasets but two, namely Micro-mass and MNIST;
    \item Considering coreset discovery as a multi-objective problem (minimize classification error and minimize coreset size), EvoCore's coresets are never dominated by the coresets uncovered by other techniques, for all considered datasets. On the contrary, they often dominate a considerable number of other solutions.
    \item In most cases, the solution provided by EvoCore was in a region of the search space far away from the solutions provided by the other techniques, thus revealing compromises unexplored by competitors.
\end{itemize}

When compared to the state-of-the-art in coreset discovery, EvoCore thus proves to be extremely effective. The main drawback of the proposed approach is represented by the longer running time. However, the problem can be mitigated by parallelizing evaluations, as EAs can easily evaluate all individuals in the same generation at the same time, provided that enough computational resources are available.

% \clearpage

% \clearpage

% \usepackage{booktabs}
% \usepackage{graphicx}
\begin{table}[!ht]
\centering
\caption{$10$-fold cross-validation results. The mean and the standard error of the mean (s.e.m.) is reported in each column. The result with the best (highest or lowest) mean value for each metric is highlighted in \textbf{bold}.}
\label{tab:results1}
\begin{center}
\begin{small}
\begin{sc}
\resizebox{0.5\textwidth}{!}{%
\begin{tabular}{@{}lllll@{}}
\toprule
micro-mass & size & test  $F_1$ & train  $F_1$ & fit time (s) \\ 
\midrule
GIGA & $299.50 \pm 0.67$ & $0.832 \pm 0.024$ & $0.995 \pm 0.002$ & $21.80 \pm 0.22$ \\
Frank-Wolfe & $323.60 \pm 0.27$ & $0.848 \pm 0.017$ & $1.000 \pm 0.000$ & $59.55 \pm 0.28$ \\
Matching Pursuit & $289.00 \pm 1.29$ & $0.850 \pm 0.022$ & $0.992 \pm 0.002$ & $61.40 \pm 0.20$ \\
Forward Stagewise & $165.60 \pm 1.36$ & $0.830 \pm 0.033$ & $0.945 \pm 0.007$ & $60.49 \pm 0.32$ \\
Ortho Pursuit & $323.00 \pm 0.00$ & \bm{$0.851 \pm 0.018$} & $1.000 \pm 0.000$ & $231.35 \pm 139.34$ \\
LAR & $323.00 \pm 0.00$ & \bm{$0.851 \pm 0.018$} & $1.000 \pm 0.000$ & $21.84 \pm 0.29$ \\
% ImportanceSampling & $324.00 \pm 0.00$ & \bm{$0.851 \pm 0.017$} & $1.000 \pm 0.000$ & \bm{$1.74 \pm 0.33$} \\
% % Random sampling & $324.00 \pm 0.00$ & \bm{$0.851 \pm 0.017$} & $1.000 \pm 0.000$ & \bm{$1.58 \pm 0.19$} \\
EvoCore & $\bm{71.60 \pm 9.54}$ & $0.839 \pm 0.023$ & $0.968 \pm 0.009$ & $227.98 \pm 2.95$ \\
\midrule
soybean & size & test  $F_1$ & train  $F_1$ & fit time (s) \\
\midrule
GIGA & $362.70 \pm 14.11$ & $0.820 \pm 0.029$ & $0.860 \pm 0.008$ & $3.56 \pm 0.12$ \\
Frank-Wolfe & $420.70 \pm 10.74$ & $0.842 \pm 0.018$ & $0.865 \pm 0.012$ & $1.79 \pm 0.10$ \\
Matching Pursuit & $342.10 \pm 13.52$ & $0.801 \pm 0.022$ & $0.829 \pm 0.011$ & $2.34 \pm 0.11$ \\
Forward Stagewise & $105.80 \pm 6.62$ & $0.737 \pm 0.024$ & $0.786 \pm 0.017$ & $1.94 \pm 0.12$ \\
Ortho Pursuit & \bm{$30.60 \pm 1.50$} & $0.624 \pm 0.033$ & $0.680 \pm 0.028$ & $0.81 \pm 0.19$ \\
LAR & $37.10 \pm 0.35$ & $0.694 \pm 0.024$ & $0.717 \pm 0.012$ & \bm{$0.80 \pm 0.16$} \\
% ImportanceSampling & $614.70 \pm 0.15$ & \bm{$0.897 \pm 0.017$} & $0.927 \pm 0.002$ & \bm{$0.65 \pm 0.08$} \\
% Random sampling & $614.70 \pm 0.15$ & $0.897 \pm 0.017$ & $0.927 \pm 0.002$ & $0.84 \pm 0.17$ \\
EvoCore & $152.80 \pm 12.95$ & \bm{$0.911 \pm 0.019$} & $0.956 \pm 0.001$ & $96.87 \pm 0.91$ \\
\midrule
credit-g & size & test  $F_1$ & train  $F_1$ & fit time (s) \\ \midrule
GIGA & $195.50 \pm 27.65$ & $0.718 \pm 0.015$ & $0.721 \pm 0.005$ & $1.19 \pm 0.18$ \\
Frank-Wolfe & $537.20 \pm 57.91$ & $0.738 \pm 0.015$ & $0.739 \pm 0.006$ & $1.93 \pm 0.07$ \\
Matching Pursuit & $398.20 \pm 79.64$ & $0.724 \pm 0.011$ & $0.737 \pm 0.006$ & $2.09 \pm 0.09$ \\
Forward Stagewise & $67.20 \pm 1.91$ & $0.671 \pm 0.019$ & $0.676 \pm 0.011$ & $2.06 \pm 0.11$ \\
Ortho Pursuit & \bm{$19.20 \pm 0.80$} & $0.668 \pm 0.015$ & $0.658 \pm 0.009$ & $0.63 \pm 0.12$ \\
LAR & $20.30 \pm 0.21$ & $0.636 \pm 0.017$ & $0.636 \pm 0.014$ & $0.67 \pm 0.10$ \\
% ImportanceSampling & $899.90 \pm 0.10$ & \bm{$0.748 \pm 0.015$} & $0.757 \pm 0.002$ & \bm{$0.74 \pm 0.13$} \\
% Random sampling & $900.00 \pm 0.00$ & \bm{$0.748 \pm 0.015$} & $0.757 \pm 0.002$ & $0.73 \pm 0.16$ \\
EvoCore & $29.60 \pm 8.28$ & \bm{$0.743 \pm 0.015$} & $0.773 \pm 0.007$ & $124.58 \pm 0.92$ \\
\midrule
kr-vs-kp & size & test  $F_1$ & train  $F_1$ & fit time (s) \\ 
\midrule
GIGA & $2227.30 \pm 32.20$ & $0.889 \pm 0.022$ & $0.939 \pm 0.002$ & $4.48 \pm 0.22$ \\
Frank-Wolfe & $2395.60 \pm 16.58$ & $0.891 \pm 0.025$ & $0.940 \pm 0.001$ & $5.18 \pm 0.11$ \\
Matching Pursuit & $2425.70 \pm 19.22$ & $0.896 \pm 0.024$ & $0.942 \pm 0.002$ & $6.25 \pm 0.17$ \\
Forward Stagewise & $297.40 \pm 24.03$ & $0.821 \pm 0.039$ & $0.878 \pm 0.011$ & $5.71 \pm 0.13$ \\
Ortho Pursuit & \bm{$28.70 \pm 2.42$} & $0.695 \pm 0.029$ & $0.732 \pm 0.014$ & \bm{$0.76 \pm 0.17$} \\
LAR & $36.10 \pm 0.18$ & $0.702 \pm 0.039$ & $0.743 \pm 0.014$ & $0.87 \pm 0.16$ \\
% ImportanceSampling & $2752.40 \pm 3.19$ & $0.892 \pm 0.024$ & $0.944 \pm 0.001$ & \bm{$0.94 \pm 0.16$} \\
% Random sampling & $2788.40 \pm 2.99$ & $0.892 \pm 0.024$ & $0.943 \pm 0.002$ & $1.10 \pm 0.30$ \\
EvoCore & $145.50 \pm 28.12$ & \bm{$0.937 \pm 0.012$} & $0.968 \pm 0.001$ & $146.55 \pm 0.53$ \\
\midrule
abalone & size & test  $F_1$ & train  $F_1$ & fit time (s) \\ 
\midrule
GIGA & $56.60 \pm 2.87$ & $0.088 \pm 0.005$ & $0.089 \pm 0.004$ & $1.69 \pm 0.25$ \\
Frank-Wolfe & $73.60 \pm 2.40$ & $0.092 \pm 0.007$ & $0.098 \pm 0.004$ & $2.96 \pm 0.11$ \\
Matching Pursuit & $61.00 \pm 2.03$ & $0.089 \pm 0.007$ & $0.095 \pm 0.002$ & $3.13 \pm 0.12$ \\
Forward Stagewise & $40.40 \pm 1.94$ & $0.083 \pm 0.010$ & $0.083 \pm 0.009$ & $3.32 \pm 0.15$ \\
Ortho Pursuit & \bm{$29.80 \pm 0.49$} & $0.092 \pm 0.006$ & $0.096 \pm 0.005$ & $1.14 \pm 0.26$ \\
LAR & \bm{$29.80 \pm 0.47$} & $0.082 \pm 0.007$ & $0.090 \pm 0.009$ & $1.61 \pm 0.61$ \\
% ImportanceSampling & $3250.40 \pm 6.18$ & \bm{$0.202 \pm 0.008$} & $0.213 \pm 0.001$ & \bm{$1.16 \pm 0.22$} \\
% Random sampling & $3498.80 \pm 5.85$ & \bm{$0.199 \pm 0.007$} & $0.211 \pm 0.001$ & \bm{$1.05 \pm 0.16$} \\
EvoCore & $232.90 \pm 32.43$ & \bm{$0.186 \pm 0.009$} & $0.194 \pm 0.003$ & $44.52 \pm 0.08$ \\
\midrule
isolet & size & test  $F_1$ & train  $F_1$ & fit time (s) \\ 
\midrule
GIGA & $1763.10 \pm 13.97$ & $0.856 \pm 0.007$ & $0.902 \pm 0.002$ & $208.60 \pm 0.35$ \\
Frank-Wolfe & $1814.50 \pm 97.73$ & $0.854 \pm 0.005$ & $0.902 \pm 0.002$ & $543.19 \pm 0.44$ \\
Matching Pursuit & $1581.20 \pm 23.01$ & $0.839 \pm 0.006$ & $0.883 \pm 0.002$ & $560.61 \pm 1.67$ \\
Forward Stagewise & $364.90 \pm 5.67$ & $0.620 \pm 0.015$ & $0.655 \pm 0.009$ & $564.02 \pm 0.62$ \\
Ortho Pursuit & \bm{$82.20 \pm 10.12$} & $0.536 \pm 0.033$ & $0.559 \pm 0.026$ & $6.58 \pm 0.89$ \\
LAR & $617.00 \pm 0.00$ & $0.614 \pm 0.008$ & $0.659 \pm 0.002$ & $433.81 \pm 40.76$ \\
% ImportanceSampling & $5286.60 \pm 5.49$ & \bm{$0.912 \pm 0.005$} & $0.957 \pm 0.001$ & \bm{$4.45 \pm 0.58$} \\
% Random sampling & $5316.30 \pm 7.00$ & \bm{$0.912 \pm 0.006$} & $0.958 \pm 0.000$ & \bm{$4.23 \pm 0.55$} \\
EvoCore & $3025.50 \pm 150.26$ & \bm{$0.905 \pm 0.006$} & $0.952 \pm 0.002$ & $7029.46 \pm 49.88$ \\
\midrule
jm1 & size & test  $F_1$ & train  $F_1$ & fit time (s) \\ 
\midrule
GIGA & $284.10 \pm 17.68$ & $0.748 \pm 0.010$ & $0.748 \pm 0.001$ & $8.33 \pm 0.52$ \\
Frank-Wolfe & $531.70 \pm 19.20$ & $0.741 \pm 0.006$ & $0.745 \pm 0.002$ & $26.58 \pm 0.13$ \\
Matching Pursuit & $287.50 \pm 13.64$ & $0.744 \pm 0.005$ & $0.748 \pm 0.002$ & $27.25 \pm 0.12$ \\
Forward Stagewise & $68.50 \pm 1.29$ & $0.743 \pm 0.005$ & $0.744 \pm 0.006$ & $27.61 \pm 0.11$ \\
Ortho Pursuit & \bm{$14.20 \pm 1.21$} & $0.739 \pm 0.006$ & $0.745 \pm 0.005$ & \bm{$0.87 \pm 0.10$} \\
LAR & $21.30 \pm 0.15$ & $0.752 \pm 0.012$ & $0.754 \pm 0.003$ & $1.11 \pm 0.13$ \\
% ImportanceSampling & $5302.10 \pm 16.06$ & $0.743 \pm 0.009$ & $0.749 \pm 0.002$ & $1.19 \pm 0.20$ \\
% Random sampling & $6252.00 \pm 10.81$ & $0.745 \pm 0.009$ & $0.750 \pm 0.002$ & $1.32 \pm 0.26$ \\
EvoCore & $54.00 \pm 14.35$ & \bm{$0.771 \pm 0.013$} & $0.788 \pm 0.002$ & $844.78 \pm 1.94$ \\
% \midrule
\bottomrule
\end{tabular}%
}
\end{sc}
\end{small}
\end{center}
\vskip -0.1in
\end{table}

% \usepackage{booktabs}
% \usepackage{graphicx}
\begin{table}[!ht]
\centering
\caption{$10$-fold cross-validation results. The mean and the standard error of the mean (s.e.m.) is reported in each column. The result with the best (highest or lowest) mean value for each metric is highlighted in \textbf{bold}.}
\label{tab:results2}
\begin{center}
\begin{small}
\begin{sc}
\resizebox{0.55\textwidth}{!}{%
\begin{tabular}{@{}lllll@{}}
\toprule
gas-drift & size & test  $F_1$ & train  $F_1$ & fit time (s) \\ 
\midrule
GIGA & $402.70 \pm 6.11$ & $0.834 \pm 0.026$ & $0.876 \pm 0.016$ & $77.82 \pm 0.48$ \\
Frank-Wolfe & $531.70 \pm 23.01$ & $0.825 \pm 0.025$ & $0.886 \pm 0.011$ & $195.76 \pm 0.21$ \\
Matching Pursuit & $324.40 \pm 8.06$ & $0.801 \pm 0.031$ & $0.859 \pm 0.017$ & $195.28 \pm 0.17$ \\
Forward Stagewise & $82.80 \pm 3.13$ & $0.710 \pm 0.034$ & $0.742 \pm 0.025$ & $195.12 \pm 0.17$ \\
Ortho Pursuit & \bm{$21.40 \pm 2.66$} & $0.513 \pm 0.046$ & $0.579 \pm 0.034$ & \bm{$1.21 \pm 0.10$} \\
LAR & $128.10 \pm 0.10$ & $0.826 \pm 0.031$ & $0.887 \pm 0.007$ & $20.83 \pm 1.15$ \\
% ImportanceSampling & $6301.50 \pm 12.40$ & \bm{$0.930 \pm 0.021$} & $0.965 \pm 0.002$ & $1.99 \pm 0.21$ \\
% Random sampling & $6876.90 \pm 8.42$ & $0.931 \pm 0.020$ & $0.971 \pm 0.002$ & $2.14 \pm 0.26$ \\
EvoCore & $813.10 \pm 162.08$ & \bm{$0.946 \pm 0.016$} & $0.986 \pm 0.001$ & $3517.78 \pm 30.70$ \\
\midrule
mozilla4 & size & test  $F_1$ & train  $F_1$ & fit time (s) \\ \midrule
GIGA & $17.30 \pm 1.17$ & $0.670 \pm 0.037$ & $0.665 \pm 0.029$ & $1.04 \pm 0.21$ \\
Frank-Wolfe & $37.90 \pm 1.31$ & $0.624 \pm 0.038$ & $0.630 \pm 0.035$ & $6.26 \pm 0.15$ \\
Matching Pursuit & $32.70 \pm 1.30$ & $0.636 \pm 0.029$ & $0.628 \pm 0.021$ & $6.46 \pm 0.06$ \\
Forward Stagewise & $15.90 \pm 1.70$ & $0.608 \pm 0.035$ & $0.578 \pm 0.024$ & $6.39 \pm 0.08$ \\
Ortho Pursuit & \bm{$5.20 \pm 0.20$} & $0.628 \pm 0.037$ & $0.604 \pm 0.044$ & \bm{$0.96 \pm 0.18$} \\
LAR & $6.70 \pm 0.40$ & $0.620 \pm 0.034$ & $0.611 \pm 0.025$ & $2.73 \pm 1.31$ \\
% ImportanceSampling & $6921.80 \pm 9.82$ & $0.792 \pm 0.009$ & $0.798 \pm 0.003$ & \bm{$1.07 \pm 0.07$} \\
% Random sampling & $7141.20 \pm 10.55$ & $0.775 \pm 0.011$ & $0.786 \pm 0.003$ & $0.99 \pm 0.07$ \\
EvoCore & $81.00 \pm 8.98$ & \bm{$0.912 \pm 0.010$} & $0.912 \pm 0.001$ & $2041.79 \pm 4.50$ \\
\midrule
letter & size & test  $F_1$ & train  $F_1$ & fit time (s) \\
\midrule
GIGA & $139.00 \pm 5.31$ & $0.213 \pm 0.007$ & $0.215 \pm 0.007$ & $1.13 \pm 0.12$ \\
Frank-Wolfe & $258.20 \pm 15.72$ & $0.244 \pm 0.008$ & $0.246 \pm 0.009$ & $38.61 \pm 0.17$ \\
Matching Pursuit & $180.50 \pm 6.80$ & $0.224 \pm 0.009$ & $0.227 \pm 0.009$ & $39.00 \pm 0.14$ \\
Forward Stagewise & $85.20 \pm 5.90$ & $0.188 \pm 0.007$ & $0.192 \pm 0.008$ & $39.17 \pm 0.12$ \\
Ortho Pursuit & \bm{$29.30 \pm 0.40$} & $0.247 \pm 0.010$ & $0.246 \pm 0.009$ & \bm{$0.95 \pm 0.11$} \\
LAR & $30.30 \pm 0.37$ & $0.238 \pm 0.007$ & $0.243 \pm 0.007$ & $8.54 \pm 5.20$ \\
% ImportanceSampling & $7535.00 \pm 7.80$ & $0.463 \pm 0.004$ & $0.464 \pm 0.003$ & \bm{$1.10 \pm 0.08$} \\
% Random sampling & $7678.30 \pm 9.60$ & $0.522 \pm 0.004$ & $0.527 \pm 0.002$ & $1.08 \pm 0.13$ \\
EvoCore & $724.30 \pm 66.09$ & \bm{$0.659 \pm 0.004$} & $0.669 \pm 0.002$ & $6916.12 \pm 33.55$ \\
\midrule
amazon-employee-access & size & test  $F_1$ & train  $F_1$ & fit time (s) \\ \midrule
GIGA & $56.70 \pm 4.33$ & $0.912 \pm 0.001$ & $0.912 \pm 0.001$ & $1.28 \pm 0.23$ \\
Frank-Wolfe & $150.00 \pm 3.96$ & $0.912 \pm 0.002$ & $0.912 \pm 0.002$ & $41.27 \pm 0.17$ \\
Matching Pursuit & $95.00 \pm 14.50$ & $0.914 \pm 0.000$ & $0.914 \pm 0.000$ & $39.43 \pm 0.21$ \\
Forward Stagewise & $45.40 \pm 4.84$ & $0.894 \pm 0.010$ & $0.894 \pm 0.010$ & $39.09 \pm 0.16$ \\
Ortho Pursuit & \bm{$9.80 \pm 0.29$} & $0.853 \pm 0.016$ & $0.852 \pm 0.015$ & \bm{$0.89 \pm 0.08$} \\
LAR & $10.20 \pm 0.33$ & $0.812 \pm 0.037$ & $0.811 \pm 0.037$ & $8.73 \pm 5.36$ \\
% ImportanceSampling & $8370.10 \pm 7.67$ & $0.914 \pm 0.000$ & $0.914 \pm 0.000$ & $1.18 \pm 0.13$ \\
% Random sampling & $8464.30 \pm 7.90$ & $0.914 \pm 0.000$ & $0.914 \pm 0.000$ & $1.14 \pm 0.08$ \\
EvoCore & $10.40 \pm 2.38$ & \bm{$0.915 \pm 0.000$} & $0.915 \pm 0.000$ & $14098.09 \pm 177.97$ \\
\midrule
electricity & size & test  $F_1$ & train  $F_1$ & fit time (s) \\ \midrule
GIGA & $81.10 \pm 9.03$ & $0.482 \pm 0.040$ & $0.513 \pm 0.020$ & $1.27 \pm 0.18$ \\
Frank-Wolfe & $190.40 \pm 16.92$ & $0.525 \pm 0.028$ & $0.553 \pm 0.014$ & $47.35 \pm 0.17$ \\
Matching Pursuit & $86.00 \pm 9.74$ & $0.552 \pm 0.046$ & $0.553 \pm 0.036$ & $47.97 \pm 0.14$ \\
Forward Stagewise & $38.20 \pm 4.29$ & $0.592 \pm 0.049$ & $0.618 \pm 0.037$ & $47.47 \pm 0.14$ \\
Ortho Pursuit & \bm{$7.90 \pm 0.10$} & $0.581 \pm 0.026$ & $0.622 \pm 0.027$ & \bm{$0.97 \pm 0.14$} \\
LAR & $8.40 \pm 0.22$ & $0.607 \pm 0.034$ & $0.607 \pm 0.024$ & $0.99 \pm 0.10$ \\
% ImportanceSampling & $8715.40 \pm 11.76$ & $0.655 \pm 0.018$ & $0.687 \pm 0.005$ & $1.32 \pm 0.10$ \\
% Random sampling & $8863.20 \pm 8.75$ & $0.692 \pm 0.028$ & $0.726 \pm 0.004$ & $1.23 \pm 0.11$ \\
EvoCore & $41.40 \pm 25.20$ & \bm{$0.693 \pm 0.029$} & $0.759 \pm 0.003$ & $38331.78 \pm 84.84$ \\
\midrule
mnist & size & test  $F_1$ & train  $F_1$ & fit time (s) \\ \midrule
GIGA & $5552.30 \pm 319.09$ & $0.797 \pm 0.003$ & $0.807 \pm 0.003$ & \bm{$1.41 \pm 0.08$} \\
Frank-Wolfe & $7306.20 \pm 240.28$ & \bm{$0.819 \pm 0.004$} & $0.826 \pm 0.002$ & $3075.35 \pm 172.95$ \\
Matching Pursuit & $6250.90 \pm 157.99$ & $0.807 \pm 0.004$ & $0.816 \pm 0.002$ & $3080.56 \pm 128.81$ \\
Forward Stagewise & $587.00 \pm 5.79$ & $0.442 \pm 0.004$ & $0.449 \pm 0.004$ & $3009.10 \pm 166.47$ \\
Ortho Pursuit & $128.80 \pm 20.96$ & $0.588 \pm 0.019$ & $0.590 \pm 0.018$ & $46.48 \pm 5.72$ \\
LAR & $711.00 \pm 0.26$ & $0.468 \pm 0.007$ & $0.477 \pm 0.005$ & $746.32 \pm 41.71$ \\
% ImportanceSampling & $9067.90 \pm 8.64$ & \bm{$0.834 \pm 0.004$} & $0.843 \pm 0.001$ & $9.35 \pm 0.83$ \\
% Random sampling & $9234.00 \pm 6.19$ & \bm{$0.835 \pm 0.004$} & $0.842 \pm 0.000$ & $10.44 \pm 0.92$ \\
EvoCore & \bm{$82.00 \pm 3.86$} & $0.772 \pm 0.006$ & $0.773 \pm 0.003$ & $4001.79 \pm 157.18$ \\
\midrule
covertype & size & test  $F_1$ & train  $F_1$ & fit time (s) \\ \midrule
GIGA & $5788.40 \pm 265.86$ & $0.583 \pm 0.027$ & $0.637 \pm 0.006$ & $1319.12 \pm 121.27$ \\
Frank-Wolfe & $4735.80 \pm 413.56$ & $0.584 \pm 0.027$ & $0.626 \pm 0.007$ & $1888.11 \pm 99.88$ \\
Matching Pursuit & $5421.30 \pm 129.91$ & $0.590 \pm 0.027$ & $0.630 \pm 0.007$ & $1893.71 \pm 96.84$ \\
Forward Stagewise & $264.20 \pm 12.45$ & $0.500 \pm 0.019$ & $0.533 \pm 0.006$ & $1975.60 \pm 46.11$ \\
Ortho Pursuit & \bm{$52.60 \pm 0.22$} & $0.486 \pm 0.016$ & $0.501 \pm 0.012$ & $17.25 \pm 1.44$ \\
LAR & $53.30 \pm 0.21$ & $0.476 \pm 0.021$ & $0.505 \pm 0.014$ & $29.26 \pm 2.52$ \\
% ImportanceSampling & $9862.40 \pm 3.88$ & $0.617 \pm 0.031$ & $0.679 \pm 0.004$ & \bm{$12.72 \pm 0.57$} \\
% Random sampling & $9902.00 \pm 3.90$ & $0.624 \pm 0.033$ & $0.679 \pm 0.003$ & \bm{$12.03 \pm 0.43$} \\
EvoCore & $328.40 \pm 59.92$ & \bm{$0.643 \pm 0.032$} & $0.701 \pm 0.003$ & $4881.47 \pm 106.46$ \\
\bottomrule
\end{tabular}%
}
\end{sc}
\end{small}
\end{center}
\vskip -0.1in
\end{table}

\section{Conclusions}
\label{sec:conclusions}
In this work, a novel alternative to coreset discovery is presented. 
% Starting from the idea that samples representative of a training set do not necessarily need to really appear inside the training set itself, an evolutionary, generative approach to create artificial coresets, archetype sets, for classification is proposed. 
Experimental results suggest that the performance of ML classifiers would not be a function of the \textit{size} of the training set, but rather a function of the \textit{mutual position} of the training samples in the feature space. 
% By exploiting the original training set and by relaxing the constraint of sample positions, the presented methodology generates a new, smaller data set suited for each classifier in order to provide the best generalization ability. 
Future works will explore the possibility of extending coreset discovery to regression and clustering.

\section*{Software and Data}

Should this paper be accepted, a link to a repository with all the code needed to reproduce the experiments will be provided.

\nocite{langley00}

\bibliography{bibliography}
\bibliographystyle{icml2020}

\end{document}